\documentclass[letterpaper]{article}
\usepackage{aaai}
\usepackage{times}
\usepackage{helvet}
\usepackage{courier}
\usepackage{latexsym}
\usepackage{ucs}
\usepackage{amsmath}
\usepackage{amsfonts}
\usepackage{amssymb}
\usepackage[american]{babel}
\usepackage{parskip}
\usepackage{graphicx}
\usepackage{algorithm}
\usepackage{algpseudocode}
\begin{document}
%
\title{Learning Scripts as Hidden Markov Models}
\author{J. Walker Orr, Prasad Tadepalli, Janardhan Rao Doppa, Xiaoli Fern, Thomas G. Dietterich \\ \{orr,tadepall,doppa,xfern,tgd\}@eecs.oregonstate.edu \\School of EECS, Oregon State Univserity, Corvallis OR 97331}
\maketitle
\begin{abstract}
\begin{quote}
Scripts have been proposed to model the stereotypical event sequences found in
narratives. They can be applied to make a variety of inferences including filling
gaps in the narratives and resolving ambiguous references. This paper proposes
the first formal framework for scripts based on Hidden Markov Models (HMMs). Our
framework supports robust inference and learning algorithms, which are lacking in
previous clustering models. We develop an algorithm for structure and parameter
learning based on Expectation Maximization and evaluate it on a number of natural
datasets. The results show that our algorithm is superior to
several informed baselines for predicting missing 
events in partial observation sequences.
%

\end{quote}
\end{abstract}

\section{Introduction}
\vspace{-.05in}
Scripts were developed as a means of representing stereotypical event sequences and interactions in narratives. 
The benefits of scripts for encoding common sense knowledge, 
filling in gaps in a story, resolving 
ambiguous references, and answering comprehension 
questions have been amply demonstrated in the early work 
in natural language understanding \cite{schank1977scripts}. 
The earliest attempts to learn scripts were based on
 explanation-based learning, which can be characterized as example-guided deduction from first principles \cite{dejong1981,dejong1986EBL}. 
While this approach is successful in generalizing from a small number of examples, it requires a strong domain theory, which limits its applicability. 

More recently, some new graph-based algorithms for inducing script-like structures from text have emerged.  ``Narrative Chains'' is a narrative model similar to Scripts \cite{chambers2008unsupervised}. Each Narrative Chain is a directed graph indicating the most frequent temporal relationship between the events in the chain. Narrative Chains are learned by a novel application of pairwise mutual information and temporal relation learning. 
Another graph learning approach employs Multiple Sequence Alignment in conjunction with a semantic similarity function to cluster sequences of event descriptions into a directed graph \cite{regneri2010learning}. More recently still, graphical models have been proposed for representing script-like knowledge, but these lack the temporal component that is central to this paper and to the early script work. These models instead focus on learning bags of related events \cite{chambers2013event,kit2013probabilistic}.

While the above approches demonstrate the learnability of script-like knowledge, they do not offer a probabilistic framework to reason robustly under uncertainty taking into account the temporal order of events. 
In this paper we present the first formal representation of scripts as Hidden Markov Models (HMMs), which support robust inference and effective learning algorithms.
The states of the HMM correspond to event types in scripts, such as entering a restaurant or opening a door. Observations correspond to natural language sentences that describe the event instances that occur in the story, e.g., ``John went to Starbucks. He came back after ten minutes.''  The standard inference algorithms, such as the Forward-Backward algorithm, are able to answer questions about the hidden states given the observed sentences, for example, ``What did John do in Starbucks?'' 

There are two complications that need to be dealt with to adapt HMMs to model narrative scripts. First, both the set of states, i.e., event types, and the set of observations are not pre-specified but are to be learned from data. We assume that the set of possible observations and the set of event types to be bounded but unknown. We employ the clustering algorithm proposed in \cite{regneri2010learning} to reduce the natural language sentences, i.e., event descriptions,  to a small set of observations and states based on their Wordnet similarity.

The second complication of narrative texts is that many events may be omitted either in the narration or by the event extraction process. 
More importantly, there is no indication of a time lapse or a gap in the story, so the standard forward-backward algorithm does not apply.  To account for this, we allow the states to skip generating observations with some probability. This kind of HMMs, with insertions and gaps, have been considered previously in speech processing \cite{bahl} and in computational biology \cite{profileHMMs}. We refine these models by allowing state-dependent missingness, without introducing additional ``insert states'' or ``delete states'' as in \cite{profileHMMs}. 
In this paper, we restrict our attention to the so-called ``Left-to-Right HMMs'' which 
have acyclic graphical structure with possible self-loops, as they support more efficient inference algorithms than general HMMs and suffice to model most of the natural scripts. 

We consider the problem of learning the structure and parameters of scripts in the form of HMMs from sequences of natural language sentences. Our solution to script learning is a novel bottom-up method for structure learning, called {\em SEM-HMM}, which is inspired by Bayesian Model Merging (BMM) \cite{stolcke1994best} and Structural Expectation Maximization (SEM)  \cite{friedman1998bayesian}. It starts with a fully enumerated HMM representation of the event sequences and incrementally merges states and deletes edges to improve the posterior probability of the structure and the parameters given the data. We compare our approach to several informed baselines on many natural datasets and show its superior performance. We believe our work represents the first formalization of scripts that supports probabilistic inference, and paves the way for robust understanding of natural language texts. 

\section{Problem Setup} 
\vspace{-.05in}

\begin{figure}
\centering
	\includegraphics[scale=.35]{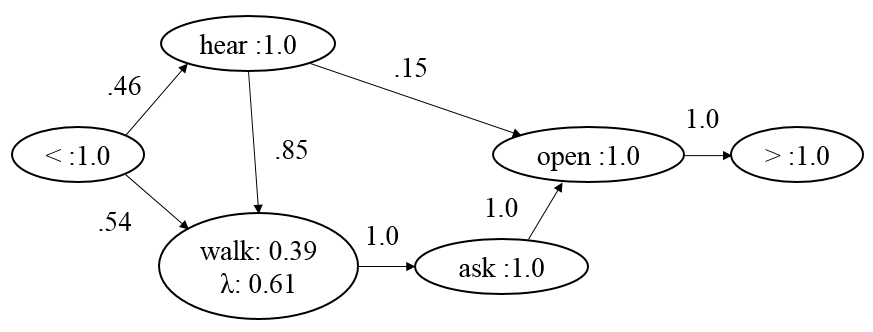}
\caption{A portion of a learned ``Answer the Doorbell'' script}
\label{fig:dbscript}
\end{figure}

Consider an activity such as answering the doorbell.  An example HMM representation of this activity is illustrated in Figure \ref{fig:dbscript}.  Each box represents a state, and the text within is a set of possible event descriptions (i.e., observations). Each event description is also marked with its conditional probability.  
Each edge represents a transition from one state to another and is annotated with its conditional probability.

In this paper, we consider a special class of HMMs with the following 
properties. 
First, we allow some observations to be missing. This is a natural phenomenon in text, where not all events are mentioned or extracted. We call these null observations and represent them with a special symbol $\lambda$.  Second, we assume that the states of the HMM can be ordered such that all transitions take place only in that order. These are called Left-to-Right HMMs in the literature \cite{rabiner,bahl}. Self-transitions of states are permitted and represent ``spurious'' observations or events with multi-time step durations. While our work can be generalized to arbitrary HMMs, we find that the Left-to-Right HMMs suffice to model scripts in our corpora.  

Formally, an HMM is a 4-tuple $(Q, T, O, \Omega)$, where $Q$ is a set of states, $T(q^\prime|q)$ is the probability of transition from $q$ to $q^\prime$, $O$ is a set of possible non-null observations, and $\Omega(o|q)$ is the probability of observing $o$ when in state $q$\footnote{$\Omega$ can be straightforwardly generalized to depend on both of the states in a state transition.},  where $o \in O \cup \{\lambda\}$, and $q_n$ is the terminal state. An HMM is Left-to-Right if the states of the HMM can be ordered from $q_0$ thru $q_n$ such that $T(q_j|q_i)$ is non-zero only if $i \leq j$. We assume that our target HMM is Left-to-Right. We index its states according to a topological ordering of the transition graph. An HMM is a generative model of a distribution over sequences of observations. For convenience w.l.o.g. we assume that each time it is ``run'' to generate a sample, the HMM starts in the same initial state $q_0$, and goes through a sequence of transitions according to $T$ until it reaches the same final state $q_n$, while emitting an observation in $O \cup \{\lambda\}$ in each state according to $\Omega$. The initial state $q_0$ and the final state $q_n$ respectively emit the distinguished observation symbols, ``$<$'' and ``$>$'' in $O$, which are emitted by no other state.
The concatenation of observations in successive states consitutes a sample of the distribution represented by the HMM. Because the null observations are removed from the generated observations, the length of the output string may be smaller than the number of state transitions. It could also be larger than the number of {\em distinct} state transitions, since we allow observations to be generated on the self transitions. Thus spurious and missing observations model insertions and deletions in the outputs of HMMs without introducing special states as in profile HMMs \cite{profileHMMs}. 



In this paper we address the following problem. Given a set of narrative texts, each of which describes a stereotypical event sequence drawn from a fixed but unknown distribution, learn the structure and parameters of a Left-to-Right HMM model that best captures the distribution of the event sequences. We evaluate the algorithm on natural datasets by how well the learned HMM can predict observations removed from the test sequences.  


\section{HMM-Script Learning}
\vspace{-.05in}

At the top level, the algorithm is input a set of documents $D$, where each document is a sequence of natural language sentences that describes the same stereotypical activity.
The output of the algorithm is a Left-to-Right HMM that represents that activity. 

Our approach has four main components, which are described in the next four subsections: Event Extraction, Parameter Estimation, Structure Learning, and Structure Scoring. The event extraction step clusters the input sentences into event types and replaces the sentences with the corresponding cluster labels.  After extraction, the event sequences are iteratively merged with the current HMM in batches of size $r$ starting with an empty HMM. Structure Learning then merges pairs of states (nodes) and removes state transitions (edges) by greedy hill climbing guided by the improvement in approximate  
posterior probability of the HMM.  Once the hill climbing converges to a local optimum, the maxmimum likelihood HMM parameters are re-estimated using the EM procedure based on all the data seen so far. Then the next batch of $r$ sequences are processed.  We will now describe these steps in more detail.

\subsection{Event Extraction}
\vspace{-.05in}

Given a set of sequences of sentences, the event extraction algorithm clusters them into events and arranges them into a tree structured HMM. For this step, we assume that each sentence has a simple structure that consists of a single verb and an object. We make the further simplifying assumption that the sequences of sentences in all documents describe the events in temporal order. Although this assumption is often violated in natural documents, we ignore this problem to focus on script learning. There have been some approaches in previous work that specifically address the problem of inferreing temporal order of events from texts, e.g., see \cite{Raghavan:temporal}.


Given the above assumptions, following \cite{regneri2010learning}, we apply a simple agglomerative clustering algorithm that uses a semantic similarity function over sentence pairs  $Sim(S_1, S_2)$ given by $w_1 PS(V_1, V_2) + w_2 PS(O_1, O_2)$, where $V_i$ is the verb and $O_i$ is the object in the sentence $S_i$. Here $PS(w, v)$ is the path similarity metric from Wordnet \cite{miller1995wordnet}. It is applied to the first verb (preferring verbs that are not stop words) and to the objects from each pair of sentences.  The constants $w_1$ and $w_2$ are tuning parameters that adjust the relative importance of each component. Like \cite{regneri2010learning}, we found that a high weight on the verb similarity was important to finding meaningful clusters of events.  
The most frequent verb in each cluster is extracted to name the event type that corresponds to that cluster. 

The initial configuration of the HMM is a Prefix Tree Acceptor,
which is constructed by starting with a single event 
sequence and then adding sequences by branching the tree
at the first place the new sequence differs from it
\cite{dupont1994search,seymore1999learning}.   
By repeating this process, an HMM that fully enumerates the data is 
constructed.  


\subsection{Parameter Estimation with EM} 
\vspace{-.05in}
\label{em}

In this section we describe our parameter estimation methods.  While parameter estimation in this kind of HMM was treated earlier in the literature \cite{rabiner,bahl}, we provide a more principled approach to estimate the state-dependent probability of $\lambda$ transitions from data without introducing special insert and delete states \cite{profileHMMs}. We assume that the structure of the Left-to-Right HMM is fixed based on the preceding structure learning step, which is described in Section~\ref{struct}.

 
The main difficulty in HMM parameter estimation is that the 
states of the HMM are not observed. The Expectation-Maximization  (EM) 
procedure (also called the Baum-Welch algorithm in HMMs) alternates between estimating the hidden states in the event sequences by running the Forward-Backward algorithm (the Expectation step) and finding the maximum likelihood estimates (the Maximization step) of the transition and observation parameters of the HMM \cite{baum1970maximization}. 
Unfortunately, because of the $\lambda$-transitions the state transitions of our HMM are not necessarily aligned with the observations. Hence we 
explicitly maintain two indices, the time index $t$ and the observation index $i$.  We define $\alpha_{q_j}(t,i)$ to be the joint probability that the HMM is in state $q_j$ at time $t$ and has made the observations $\vec{o}_{0,i}$. This is computed by the forward pass of the algorithm using the following recursion. Equations \ref{eq:fw1} and \ref{eq:fw2} represent the base case of the recursion, while Equation \ref{eq:fw3} represents the case for null observations. Note that the observation index $i$ of the recursive call is not advanced unlike in the second half of Equation \ref{eq:fw3} where it is advanced for a normal observation.  We exploit the fact that the HMM is Left-to-Right and only consider transitions to $j$ from states with indices $k \leq j$. The time index $t$ is incremented starting $0$, and the observation index $i$ varies from $0$ thru $m$. 
{\footnotesize
\vspace{-.1in}
\begin{align}
&\alpha_{q_0}(0,0) = 1 \label{eq:fw1}\\
&\forall j > 0, \alpha_{q_j}(0,0) = 0 \label{eq:fw2}\\
&\alpha_{q_j}(t,i) = \sum_{0 \leq k \leq j} T(q_j|q_k) \{\Omega(\lambda|q_j) \alpha_{q_k}(t-1,i) \label{eq:fw3}\\ 
&\qquad + \Omega(o_{i}|q_j) \alpha_{q_k}(t-1,i-1) \} \nonumber
\end{align}
\vspace{-.2in}
}%

The backward part of the standard Forward-Backward algorithm starts from the last time step $\tau$ and reasons backwards. Unfortunately in our setting, we do not know $\tau$---the true number of state transitions---as some of the observations are missing.  Hence, we define $\beta_{q_j}(t,i)$ as the conditional probability of observing $\vec{o}_{i+1,m}$ in the remaining $t$ steps given that the current state is $q_j$.  This allows us to increment $t$ starting from $0$ as recursion proceeds, rather than decrementing it from $\tau$.
{\footnotesize
\vspace{-.1in}
\begin{align}
&\beta_{q_n}(0,m) = 1 \\
&\forall j < n, \beta_{q_j}(0,m) = 0 \\
&\beta_{q_j}(t,i) = \sum_{j \leq k} T(q_k|q_j) \{ \Omega(\lambda|q_k) \beta_{q_k}(t-1,i) \\  
&\qquad + \Omega(o_{i+1}|q_k) \beta_{q_k}(t-1,i+1) \} \nonumber 
\end{align}
\vspace{-.2in}
}%

Equation \ref{eq:z} calculates the probability of the observation sequence $z = P(\vec{o})$, which is computed by marginalizing $\alpha_q(t,m)$ over time $t$ and state $q$ and setting the second index $i$ to the length of the observation sequence $m$. The quantity $z$ serves as the normalizing factor for the last three equations.
{\footnotesize
\begin{align}
&z = P(\vec{o}) = \sum_{q \in Q} \sum_{t} \alpha_{q}(t,m) \label{eq:z}\\
&\gamma_q(t,i) = P(q | \vec{o}) = z^{-1} \sum_{\tau} \alpha_q(t,i) \beta_q(\tau-t,i) \label{eq:gamma}\\
&\delta_{q,q^\prime \uparrow \lambda}(t) = P(q \rightarrow q^\prime,\lambda | \vec{o}) = z^{-1}T(q^\prime | q) \Omega(\lambda |q^\prime) \label{eq:delta1}\\
&\qquad \sum_{\tau} \sum_i \{ \alpha_q(t,i) \beta_{q^\prime}(\tau-t-1,i) \} \nonumber\\
&\forall o \in \Omega, \delta_{q,q^\prime \uparrow o}(t) = P(q \rightarrow q^\prime,o | \vec{o}) \label{eq:delta2}\\
&\qquad = z^{-1}T(q^\prime | q) \Omega(o | q^\prime) \nonumber\\
&\qquad \sum_{\tau} \sum_i \{ \alpha_q(t,i) I(o_{i+1}=o) \beta_{q^\prime}(\tau-t-1,i+1) \} \nonumber
\end{align}
\vspace{-.1in}
}%

Equation \ref{eq:gamma}, the joint distribution of the state and observation index $\gamma$ at time $t$ is computed by convolution, i.e., multiplying the $\alpha$ and $\beta$ that correspond to the same time step and the same state and marginalizing out the length of the state-sequence $\tau$. Convolution is necessary, as the length of the state-sequence $\tau$ is a random variable equal to the sum of the corresponding time indices of $\alpha$ and $\beta$.

Equation \ref{eq:delta1} computes the joint probability of a state-transition associated with a null observation by first multiplying the state transition probability by the null observation probability given the state transition and the appropriate $\alpha$ and $\beta$ values. It then marginalizes out the observation index $i$. Again we need to compute a convolution with respect to $\tau$ to take into account the variation over the total number of state transitions.  
Equation \ref{eq:delta2} calculates the same probability for a non-null observation $o$. This equation is similar to Equation \ref{eq:delta1} with two differences.  First, we ensure that the observation is consistent with $o$ by multiplying the product with the indicator function $I(o_{i+1} = o)$ which is $1$ if $o_{i+1} = o$ and $0$ otherwise.  Second, we advance the observation index $i$ in the $\beta$ function.

Since the equations above are applied to each individual observation sequence, $\alpha$, $\beta$, $\gamma$, and $\delta$ all have an implicit index $s$ which denotes the observation sequence and has been omitted in the above equations. We will make it explicit below and calculate the expected counts of state visits, state transitions, and state transition observation triples. 
{\footnotesize
\begin{align}
&\forall q \in Q, C(q) = \sum_{s,t,i} \gamma_q(s,t,i) \label{eq:count}\\
&\forall q, q^\prime \in Q, C(q \rightarrow q^\prime) = \sum_{s,t,o \in \Omega \bigcup \{\lambda\}} \delta_{q,q^\prime \uparrow o}(s,t) \label{eq:transCount}\\
&\forall q, q^\prime \in Q, o \in \Omega \bigcup \{\lambda\}, \label{eq:sigCount}\\
&\qquad C(q, q^\prime \uparrow o) = \sum_{s,t} \delta_{q,q^\prime \uparrow o}(s,t) \nonumber
\end{align}
\vspace{-.2in}
}%



Equation \ref{eq:count} counts the total expected number of visits of each state in the data. Also, Equation \ref{eq:transCount} estimates the expected number of transitions between each state pair. Finally, Equation \ref{eq:sigCount} computes the expected number of observations and state-transitions including null transitions. This concludes the E-step of the EM procedure. 


The M-step of the EM procedure consists of 
Maximum Aposteriori (MAP) estimation of the 
transition and observation distributions is done 
assuming an uninformative Dirichlet prior. This 
amounts to 
adding a pseudocount of 1 to each of the next states and observation 
symbols. 
The observation distributions for the initial and final states $q_0$ and 
$q_n$ are fixed to be the Kronecker delta distributions at their true values. 
{\footnotesize
\begin{align}
&\hat{T}(q^\prime | q) = \frac{ C(q \rightarrow q^\prime) + 1}{[C(q)+ \sum_{p^{\prime} \in Q} 1]} \\
&\hat{\Omega}(o|q^\prime) = \frac {{\sum_{q} C(q, q^\prime \uparrow o)} + 1} {\sum_{o^\prime} \{ \sum_q C(q,q^\prime \uparrow o^\prime)\} + 1} 
\end{align}
\vspace{-.1in}
}%

The E-step and the M-step are repeated until convergence of the 
parameter estimates.

\subsection{Structure Learning}
\vspace{-.05in}
\label{struct}

We now describe our structure learning algorithm, SEM-HMM.  Our algorithm is inspired by Bayesian Model Merging (BMM) \cite{stolcke1994best} and Structural EM (SEM) \cite{friedman1998bayesian} and adapts them to learning HMMs with missing observations.
SEM-HMM performs a greedy hill climbing search through the space of acyclic 
HMM structures. It iteratively proposes changes to the structure either 
by merging states or by deleting edges. It evaluates each change and makes the one with the best score. 
An exact implementation of this method is expensive, because, each time a structure change is considered, the MAP parameters of the structure given the data must be re-estimated.
One of the key insights of both SEM and BMM is that this expensive 
re-estimation can be avoided in factored models by incrementally 
computing the changes to various expected counts using only local 
information. While this calculation is only approximate, it is highly 
efficient. 


During the structure search, the algorithm considers every possible structure change, i.e., merging of pairs of states and deletion of state-transitions, checks that the change does not create cycles, evaluates it according to the scoring function and selects the best scoring structure. This is repeated until the structure can no longer be improved (see Algorithm~\ref{learningAlg}).

\begin{algorithm}
\footnotesize
\begin{algorithmic}
	\Procedure{Learn}{Model $M$, Data $D$, Changes $S$}

		\While {$Not Converged$}
                        \State ${\cal M}$ = AcyclicityFilter $(S(M))$ 
			\State $M^* = argmax_{M^\prime \in {\cal M}} P(M^\prime|D)$

			\If {$P(M^* | D) \le P(M | D)$}
				\State \Return $M$
			\Else
				\State $M = M^*$
			\EndIf

		\EndWhile
	\EndProcedure
\end{algorithmic}
\caption{}
\label{learningAlg}
\end{algorithm}

The Merge States operator creates a new state from the union of a state pair's transition and observation distributions.  
It must assign transition and observation distributions to the new merged 
state. To be exact, we need to redo the parameter estimation for the changed 
structure. To compute the impact of several proposed changes efficiently,
we assume that all probabilistic state transitions and trajectories for 
the observed sequences remain the same as before except in the changed 
parts of the structure. We call this ``locality of change'' assumption, which 
allows us to 
add the corresponding expected counts from the states being merged as shown below. 
{\footnotesize
\begin{align}
&C(r) = C(p) + C(q) \nonumber\\
&C(r \rightarrow s) = C(p \rightarrow s) +  C(q \rightarrow s) \nonumber \\
&C(s \rightarrow r) = C(s \rightarrow p) +  C(s \rightarrow q) \nonumber \\
&C(r,s \uparrow o) = C(p,s \uparrow o) +  C(q,s \uparrow o) \nonumber \\
&C(s,r \uparrow o) = C(s,p \uparrow o) +  C(s,q \uparrow o) \nonumber 
\end{align}
\vspace{-.2in}
}%

The second kind of structure change we consider is edge deletion and consists of removing a transition between two states and 
redistributing its evidence along the other paths between the same states. 
Again, making the locality of change assumption, we only recompute 
the parameters of the transition and observation distributions 
that occur in the paths between the two states.
We re-estimate the parameters due to deleting an edge $(q_s,q_e)$,
by effectively redistributing the expected transitions from 
$q_s$ to $q_e$, $C(q_s \rightarrow q_e)$,
among other edges between $q_s$ and $q_e$ based on the parameters of the current model. 

This is done efficiently using a procedure similar to the 
Forward-Backward algorithm under the null observation sequence.
Algorithm~\ref{deleteedge} takes the current model $M$, an edge 
($q_s \rightarrow q_e$), and the expected count of the number of transitions 
from $q_s$ to $q_e$, $N = C(q_s \rightarrow q_e)$,  
as inputs. It updates the counts of 
the other transitions to compensate for removing the edge between 
$q_s$ and $q_e$. It initializes the $\alpha$ of $q_s$ and the $\beta$ 
of $q_e$ with 1 and the rest of the $\alpha$s and $\beta$s to $0$. 
It makes two passes through the HMM, 
first in the topological order of the nodes in the graph and the second 
in the reverse topological order. In the first, ``forward'' pass
from $q_s$ to $q_e$, 
it calculates  the $\alpha$ value of each node $q_i$ 
that represents the probability that a sequence that 
passes through $q_s$ also passes through $q_i$ 
while emitting no observation. In the second, ``backward'' pass,
it computes the $\beta$ value of a node $q_i$ 
that represents the probability that a sequence that passes
through 
$q_i$ emits no observation and later 
passes through $q_e$. The product of 
$\alpha(q_i)$ and $\beta(q_i)$ gives the probability that 
$q_i$ is passed through when going from $q_s$ to $q_t$ and emits
no observation. Multiplying it by the 
expected number of transitions $N$ gives the 
expected number of additional counts 
which are added to $C(q_i)$ 
to compensate for the deleted transition $(q_s \rightarrow q_e)$.
After the distribution of the evidence, 
all the transition and observation probabilities are re-estimated for
the nodes and edges affected by the edge deletion 

\begin{algorithm}
\footnotesize
\begin{algorithmic}

\Procedure{DeleteEdge}{Model $M$, edge $(q_s \rightarrow q_e)$, count $N$}

\State $\forall i s.t. s \leq i \leq e, \alpha(q_i) = \beta(q_i) = 0$
%
   \State $\alpha(q_s) = \beta(q_e) = 1$
\For {$i = s+1$ {\bf to}  $e$}
    \ForAll {$q_p \in Parents(q_i)$} 
      \State  $\alpha(q_p \rightarrow q_i) = \alpha(q_p) T(q_i|q_p) \Omega(\lambda|q_i)$
      \State $\alpha(q_i) = \alpha(q_i) + \alpha(q_p \rightarrow q_i)$
\EndFor
\EndFor

\For {$i = e-1$ {\bf downto} $s$} 
   \ForAll {$q_c \in Children(q_i)$} 
      \State  $\beta(q_i \rightarrow q_c) = \beta(q_c) T(q_c|q_i) \Omega(\lambda|q_c)$
      \State  $C(q_i \rightarrow q_c)$ = $C(q_i \rightarrow q_c) + \alpha(q_i \rightarrow q_c)  \beta(q_i \rightarrow q_c) N$
      \State  $C(q_i)$ = $C(q_i) + C(q_i \rightarrow q_c)$
      \State $\beta(q_i) = \beta(q_i) + \beta(q_i \rightarrow q_c)$
\EndFor
\EndFor
\EndProcedure

\end{algorithmic}
\label{deleteedge}
\caption{Forward-Backward algorithm to delete an edge and re-distribute the expected counts.}
\end{algorithm}



In principle, one could continue making incremental structural changes and 
parameter updates and never run EM again. This is exactly 
what is done in 
Bayesian Model Merging (BMM) \cite{stolcke1994best}. 
However, a series of structural changes followed 
by approximate incremental parameter updates could lead to bad 
local optima. Hence, after merging each batch of $r$ sequences into the HMM,  we
re-run EM for parameter estimation on all sequences seen thus far. 









\subsection{Structure Scoring}
\vspace{-.05in}
\label{sec:fb}

We now describe how we score the structures produced by our algorithm to select the best structure. 
We employ a Bayesian scoring function, which is 
the posterior probability of the model given the data, denoted $P(M|D)$.  The score is decomposed via Bayes Rule (i.e., $P(M|D) \propto P(M) P(D|M))$), and the denominator is omitted since it is invariant with regards to the model. 

Since each observation sequence is independent of the others, 
the data likelihood $P(D|M) = \Pi_{\vec{o} \in D} P(\vec{o})$ is calculated 
using the Forward-Backward algorithm and Equation \ref{eq:z} in
Section~\ref{em}. Because the initial model fully enumerates the data, any merge can only reduce the data likelihood. Hence, the model prior $P(M)$ must be designed to encourage generalization via state merges and edge deletions (described in Section~\ref{struct}). We employed a prior with three components: the first two components are syntactic and penalize the number of states $|Q|$ and the number of non-zero transitions $|T|$ respectively. The third component penalizes the number of frequently-observed semantic constraint violations $|C|$. In particular, the prior probabilty of the model  $P(M) = \frac{1}{Z}\exp( -(\kappa_q |Q| + \kappa_t |T| +\kappa_c |C|))$.
The $\kappa$ parameters assign weights to each component in the prior. 


The semantic constraints are learned from the event sequences for use in the model prior.  The constraints take the simple form ``$X$ never follows $Y$.''  They are learned by generating all possible such rules using 
pairwise permutations of event types, and evaluating them on the training data. 
In particular, the number of times each rule is violated is counted and a $z$-test is performed to determine if the violation rate is lower than a predetermined error rate. Those rules that pass the hypothesis test with a threshold of $0.01$ are included.  When evaluating a model, these contraints are considered violated if the model could generate a sequence of observations that violates the constraint. 



Also, in addition to incrementally computing the transition and observation counts, $C(r \rightarrow s)$ and $C(r,s \uparrow o)$, the likelihood, $P(D|M)$ can be incrementally updated with structure changes as well. Note that the likelihood can be expressed as $P(D|M) = \prod_{q, r \in Q} \prod_{o \in O} T(r|q)^{C(q \rightarrow r)} \Omega(o|r)^{C(q,r \uparrow o)}$ when the state transitions are observed. Since the state transitions are not actually observed, we approximate the above expression by replacing the observed counts with expected counts. Further, the locality of change assumption allows us to easily 
calculate the effect of changed expected counts and parameters on the likelihood by dividing it by the old products and multiplying by the new products. We call this version of our algorithm SEM-HMM-Approx.

\section{Experiments and Results}
\vspace{-.05in}

We now present our experimental results on SEM-HMM and SEM-HMM-Approx. The
evaluation task is to predict missing events from an observed sequence of events.  
For comparison, four baselines were also evaluated.  The ``Frequency'' baseline predicts the most frequent event in the training set that is not found in the observed test sequence. The ``Conditional'' baseline predicts the next event based on what most frequently follows the prior event.  
A third baseline, referred to as ``BMM,'' is a version of our algorithm that does not use EM for parameter estimation and instead only incrementally updates the parameters starting from the raw document counts. Further, it learns a standard HMM, that is, with no $\lambda$ transitions. 
This is very similar to the Bayesian Model Merging approach for HMMs \cite{stolcke1994best}. 
The fourth baseline is the same as above, but uses our EM algorithm for parameter estimation without $\lambda$ transitions. It is referred to as ``BMM + EM.'' 

\begin{table}
\footnotesize
\begin{center}
\begin{tabular}{|l|r|r|r|}
\hline
Batch Size $r$ & 2 & 5 & 10\\
\hline
SEM-HMM & 42.2\% & 45.1\% & 46.0\%\\
SEM-HMM Approx.& 43.3\% & 43.5\% & 44.3\%\\
BMM + EM & 41.1\% & 41.2\% & 42.1\%\\
BMM & 41.0\% & 39.5\% & 39.1\%\\
\hline
Conditional & \multicolumn{3}{r|}{36.2\%}\\
Frequency & \multicolumn{3}{r|}{27.3\%}\\
\hline
\end{tabular}
\caption{The average accuracy on the OMICS domains}
\end{center}
\label{table:omics}
\end{table}

The Open Minds Indoor Common Sense (OMICS) corpus was developed by the Honda Research Institute and is based upon the Open Mind Common Sense project \cite{gupta2004common}. It describes 175 common household tasks with each task having 14 to 122 narratives describing, in short sentences, the necessary steps to complete it.  Each narrative consists of temporally ordered, simple sentences from a single author that describe a plan to accomplish a task.  Examples from the ``Answer the Doorbell'' task can be found in Table 2. The OMICS corpus has 9044 individual narratives and its short and relatively consistent language lends itself to relatively easy event extraction. 


\begin{table}
\footnotesize
\begin{center}
\begin{tabular}{| l | l |}
\hline
Example 1 & Example 2 \\
\hline
\underline{Hear} the doorbell. & \underline{Listen} for the doorbell. \\
\underline{Walk} to the door. & \underline{Go} towards the door. \\
\underline{Open} the door. & \underline{Open} the door. \\
\underline{Allow} the people in. & \underline{Greet} the vistor. \\
\underline{Close} the door. & \underline{See} what the visitor wants. \\
& \underline{Say} goodbye to the visitor. \\
& \underline{Close} the door. \\
\hline
\end{tabular}
\caption{Examples from the OMICS ``Answer the Doorbell'' task with event triggers underlined}
\end{center}
\label{table:omicsexample}
\end{table}

The 84 domains with at least 50 narratives and 3 event types were used for evaluation.
For each domain, forty percent of the narratives were withheld for testing, each with one randomly-chosen event omitted.  The model was evaluated on the proportion of correctly predicted events given the remaining sequence.
On average each domain has 21.7 event types with a standard deviation of 4.6.  Further, the average narrative length across domains is 3.8 with standard deviation of 1.7.  This implies that 
only a frcation of the event types are present in any given narrative.  
There is a high degree of omission of events and many different ways of accomplishing each task.
Hence, the prediction task is reasonably difficult, as evidenced by the simple baselines.  
Neither the frequency of events nor simple temporal structure is enough to accurately fill in the gaps which indicates that most sophisticated modeling such as SEM-HMM is needed.

The average accuracy across the 84 domains for each method is found in Table 1. 
On average our method significantly out-performed all the baselines, with the average improvement in accuracy across OMICS tasks between SEM-HMM and each baseline being statistically significant at a .01 level across all pairs and on sizes of $r = 5$ and $r= 10$ using one-sided paired t-tests.  For $r=2$ improvement was not statistically greater than zero. 
We see that the results improve with batch size $r$ until $r=10$ for 
SEM-HMM and BMM+EM, but they decrease with batch size for BMM without EM.
Both of the methods which use EM depend on statistics to be robust and hence need a larger $r$ value to be accurate.  However for BMM, a smaller $r$ size means it reconciles a couple of documents with the current model in each iteration which ultimately helps guide the structure search. 
The accuracy for ``SEM-HMM Approx.'' is close to the exact version at each batch level, while only taking half the time on average.  

\section{Conclusions}
\vspace{-.05in}

In this paper, we have given the first formal treatment of scripts as 
HMMs with missing observations.  
We adapted the HMM inference and parameter estimation procedures 
to scripts and developed a new structure 
learning algorithm, SEM-HMM, based on the EM procedure. 
It improves upon BMM by allowing for $\lambda$ transitions and by
incorporating maximum likelihood parameter estimation via EM. 
We showed that our algorithm is effective in learning scripts from 
documents and performs better than 
other baselines on sequence prediction tasks. Thanks to the 
assumption of missing observations, the graphical 
structure of the scripts is usually sparse and intuitive. 
Future work includes learning from more natural text such as 
newspaper articles, enriching the representations to include objects and 
relations, and integrating HMM inference into text understanding. 



\section*{Acknowledgments}
We would like to thank Nate Chambers, Frank Ferraro, and Ben Van Durme for their helpful comments, criticism, and feedback.  Also we would like to thank the SCALE 2013 workshop. This work was supported by the DARPA and AFRL under contract No. FA8750-13-2-0033. Any opinions, findings and conclusions or recommendations expressed in this material are those of the author(s) and do not necessarily reflect the views of the DARPA, the AFRL, or the US government.


\bibliography{hmm}

\begin{thebibliography}{}

\bibitem[\protect\citeauthoryear{Bahl, Jelinek, and Mercer}{1983}]{bahl}
Bahl, L.~R.; Jelinek, F.; and Mercer, R.~L.
\newblock 1983.
\newblock A maximum likelihood approach to continuos speech recognition.
\newblock {\em IEEE Transactions in Pattern Analysis and Machine Intelligence
  (PAMI)} 5(2):179--190.

\bibitem[\protect\citeauthoryear{Baum \bgroup et al\mbox.\egroup
  }{1970}]{baum1970maximization}
Baum, L.~E.; Petrie, T.; Soules, G.; and Weiss, N.
\newblock 1970.
\newblock A maximization technique occurring in the statistical analysis of
  probabilistic functions of markov chains.
\newblock {\em The {A}nnals of {M}athematical {S}tatistics} 41(1):164--171.

\bibitem[\protect\citeauthoryear{Chambers and
  Jurafsky}{2008}]{chambers2008unsupervised}
Chambers, N., and Jurafsky, D.
\newblock 2008.
\newblock Unsupervised learning of narrative event chains.
\newblock In {\em Proceedings of the 46th Annual Meeting of the Association for
  Computational Linguistics (ACL)},  789--797.

\bibitem[\protect\citeauthoryear{Chambers}{2013}]{chambers2013event}
Chambers, N.
\newblock 2013.
\newblock Event schema induction with a probabilistic entity-driven model.
\newblock In {\em Proceedings of the 2013 Conference on Empirical Methods in
  Natural Language Processing},  1797--1807.

\bibitem[\protect\citeauthoryear{DeJong and Mooney}{1986}]{dejong1986EBL}
DeJong, G., and Mooney, R.
\newblock 1986.
\newblock Explanation-based learning: An alternative view.
\newblock {\em Machine learning} 1(2):145--176.

\bibitem[\protect\citeauthoryear{DeJong}{1981}]{dejong1981}
DeJong, G.
\newblock 1981.
\newblock Generalizations based on explanations.
\newblock In {\em Proceedings of the Seventh International Joint Conference on
  Artificial Intelligence (IJCAI)},  67--69.

\bibitem[\protect\citeauthoryear{Dupont, Miclet, and
  Vidal}{1994}]{dupont1994search}
Dupont, P.; Miclet, L.; and Vidal, E.
\newblock 1994.
\newblock What is the search space of the regular inference?
\newblock In {\em Grammatical Inference and Applications}. Springer.
\newblock  25--37.

\bibitem[\protect\citeauthoryear{Friedman}{1998}]{friedman1998bayesian}
Friedman, N.
\newblock 1998.
\newblock The {B}ayesian structural {EM} algorithm.
\newblock In {\em Proceedings of the Fourteenth conference on Uncertainty in
  artificial intelligence},  129--138.
\newblock Morgan Kaufmann Publishers Inc.

\bibitem[\protect\citeauthoryear{Gupta and
  Kochenderfer}{2004}]{gupta2004common}
Gupta, R., and Kochenderfer, M.~J.
\newblock 2004.
\newblock Common sense data acquisition for indoor mobile robots.
\newblock In {\em AAAI},  605--610.

\bibitem[\protect\citeauthoryear{Kit~Cheung, Poon, and
  Vanderwende}{2013}]{kit2013probabilistic}
Kit~Cheung, J.~C.; Poon, H.; and Vanderwende, L.
\newblock 2013.
\newblock Probabilistic frame induction.
\newblock In {\em Proceedings of Conference of the North American Chapter of
  the Association for Computational Linguistics: Human Language Technologies
  (NAACL:HLT)},  837--846.

\bibitem[\protect\citeauthoryear{Krogh \bgroup et al\mbox.\egroup
  }{1994}]{profileHMMs}
Krogh, A.; Brown, M.; Mian, I.~S.; Sjolander, K.; and Haussler, D.
\newblock 1994.
\newblock Hidden markov models in computational biology.
\newblock {\em Journal of Molecular Biology}  1501--1531.

\bibitem[\protect\citeauthoryear{Miller}{1995}]{miller1995wordnet}
Miller, G.~A.
\newblock 1995.
\newblock {WordNet:} a lexical database for english.
\newblock {\em Communications of the ACM} 38(11):39--41.

\bibitem[\protect\citeauthoryear{Rabiner}{1990}]{rabiner}
Rabiner, L.~R.
\newblock 1990.
\newblock A tutorial on hidden {M}arkov models and selected applications in
  speech recognition.
\newblock  267--296.

\bibitem[\protect\citeauthoryear{Raghavan, Fosler-Lussier, and
  Lai}{2012}]{Raghavan:temporal}
Raghavan, P.; Fosler-Lussier, E.; and Lai, A.~M.
\newblock 2012.
\newblock Learning to temporally order medical events in clinical text.
\newblock In {\em Proceedings of the 46th Annual Meeting of the Association for
  Computational Linguistics (ACL)},  70--74.

\bibitem[\protect\citeauthoryear{Regneri, Koller, and
  Pinkal}{2010}]{regneri2010learning}
Regneri, M.; Koller, A.; and Pinkal, M.
\newblock 2010.
\newblock Learning script knowledge with web experiments.
\newblock In {\em Proceedings of the 48th Annual Meeting of the Association for
  Computational Linguistics},  979--988.
\newblock Association for Computational Linguistics.

\bibitem[\protect\citeauthoryear{Schank and Abelson}{1977}]{schank1977scripts}
Schank, R., and Abelson, R.
\newblock 1977.
\newblock {\em Scripts, plans, goals and understanding: An inquiry into human
  knowledge structures.}
\newblock Lawrence Erlbaum Publishers.

\bibitem[\protect\citeauthoryear{Seymore, McCallum, and
  Rosenfeld}{1999}]{seymore1999learning}
Seymore, K.; McCallum, A.; and Rosenfeld, R.
\newblock 1999.
\newblock Learning hidden {Markov} model structure for information extraction.
\newblock In {\em AAAI Workshop on Machine Learning for Information
  Extraction},  37--42.

\bibitem[\protect\citeauthoryear{Stolcke and Omohundro}{1994}]{stolcke1994best}
Stolcke, A., and Omohundro, S.~M.
\newblock 1994.
\newblock Best-first model merging for hidden {Markov} model induction.
\newblock {\em arXiv preprint cmp-lg/9405017}.

\end{thebibliography}
\bibliographystyle{aaai}

\end{document}